\documentclass[reprint,
 amsmath,amssymb,
 aps,
]{revtex4-2}

\usepackage{subfiles}
\usepackage{graphicx}
\def\figwidth{\columnwidth}
\usepackage{dcolumn}
\usepackage{bm}
\usepackage{comment}
\usepackage{xcolor}
\newcommand{\new}[1]{#1}

\begin{document}

\preprint{APS/123-QED}

\title{A transparent approach to data representation}

\author{Sean Deyo}
\email{sjd257@cornell.edu}
\author{Veit Elser}
\affiliation{Department of Physics, Cornell University\\
Ithaca, NY 14853}

\date{\today}

\begin{abstract}
We use a binary attribute representation (BAR) model to describe a data set of Netflix viewers' ratings of movies. We classify the viewers with discrete bits rather than continuous parameters, which makes the representation compact and transparent. The attributes are easy to interpret, and we need far fewer attributes than similar methods do to achieve the same level of error. We also take advantage of the nonuniform distribution of ratings among the movies in the data set to train on a small selection of movies without compromising performance on the rest of the movies.
\end{abstract}

\maketitle


\section{Introduction}

In 2006 Netflix released a data set --- roughly 100 million ratings of 17770 titles, given by 480189 viewers --- and posed a challenge: Use this training data to predict the ratings in a separate, hidden set of ratings involving the same movies and viewers. The first to do so with a root-mean-square prediction error (RMSE) at least $10\%$ lower than that of Netflix's own system would receive a million-dollar prize \cite{feuerverger2012statistical}. Though the competition ended in 2009, when someone beat the target RMSE, the data set is still of interest for a variety of reasons. It is certainly well-studied, with more than fifty-thousand competition-entrants trying a variety of approaches to nab the prize money. The nature of the data is also intriguing: subjective and probably not ``repeatable'' because the ratings were made by humans, and discrete in a somewhat arbitrary way (1-5 stars, though the contest allowed entrants to use continuous values for the predicted ratings). In fact, it is speculated that the target RMSE that Netflix set and that the winners ultimately achieved may be just about the best possible \cite{feuerverger2012statistical}. And, of course, the fact that the data set is full of popular (and also more esoteric) films makes it culturally relevant and enjoyable to explore.

The winning submission was a ``blend'' of hundreds of different prediction methods constructed by three groups that joined forces \cite{koren2009bellkor,piotte2009pragmatic,toscher2009bigchaos}. The methods in the blend included linear algebra techniques like singular value decomposition, simple neural networks, and more complicated models involving scores of fine-tuned meta-parameters. We see this as a key early moment in the trend of neural networks away from interpretability and toward the black box paradigm. 
The increasingly complicated models and the way they produced a much better RMSE when blended together with just the right weights foreshadowed the way today's neural networks rely on greater depth and larger numbers of weights to reduce their loss. So it is fitting that we revisit the Netflix prize as a setting to explore the question of how well a single model can do with tight limits on its size and transparency.

We take inspiration from the non-negative matrix factorization (NMF) problem. In NMF, one large $m\times n$ matrix $M$ with non-negative values is factored as a product of two smaller non-negative matrices $R$ and $C$ of size $m\times l$ and $l\times n$, respectively (where $l\ll m,n$). Imagining the set of ratings as the $M$ matrix, with each row corresponding to a viewer and each column corresponding to a movie, one can think of each row of $R$ as an attribute vector for the corresponding viewer. The elements of the attribute vector represent what attributes that viewer possesses and to what degree. One element might represent how much a viewer ``likes action.'' Similarly, one can think of each column of $C$ as a weight vector for the corresponding movie, whose elements represent how well each movie satisfies the given attributes. The weight corresponding to the ``likes action'' attribute would represent how much action is in the movie. It turns out to be easier to subtract out the average ratings of each column and row of $M$, so we will not maintain the notion of non-negativity, but we preserve the most important aspect --- that the number of attributes $l$ is much smaller than the number of movies or of viewers. We will also restrict the matrix of attribute vectors $R$ to be binary --- a viewer either has an attribute or does not --- which makes the description of the viewers especially compact.

Of the many models employed in pursuit of the million-dollar prize, ours is most similar to a restricted Boltzmann machine (RBM). The RBM is a simple neural network that, in this context, assigns ``features'' to each viewer and learns weights to connect each feature for each viewer to each of the five possible ratings (1-5) \cite{salakhutdinov2007rbm}. Our approach\new{, the binary attribute representation (BAR)}, differs in that we embrace the natural ordering between the five rating options, yielding a model with fewer weights, and that we use a projection-based iteration scheme instead of the ubiquitous gradient descent/ascent. This scheme is especially important given that we are trying to infer discrete attribute bits. We shall see that \new{BAR} can achieve an RMSE lower than that of a typical RBM \cite{salakhutdinov2007rbm} and comparable to that of the individual ingredients of the prize-winning blend, all with far fewer features/attributes than these methods.

Another of our objectives is to take advantage of the nonuniformity of the data. Most viewers have not seen most movies, so there are many unconstrained elements in the matrix. But these elements are not distributed uniformly: Some movies have been rated by many viewers, others have not.
We will demonstrate that a small number of movies can be sufficient to learn the viewers' attributes, and that it is not difficult to explain the ratings of the rest of the movies with those attributes.

The competition is over, so there is no reason for us to seek the same goal as the original challenge. We treat this as a data \new{representation} or compression problem, rather than a prediction problem. Our goal is to find a compact and interpretable representation of viewers' preferences and movies' qualities. The ability to make ``predictions'' is a secondary ability that falls out of this representation for free, but optimizing these predictions is not our focus.

\section{Method}

\new{\subsection{Binary attribute representation}}
There are many viewers --- nearly half a million. We want to encode their preferences as compactly as possible. The most compact kind of information we can ask for is a set of binary values, which we will call \textit{attribute bits}. One can think of each bit as representing a preference that the viewer has --- whether the viewer ``likes action,'' for instance. Viewers that like action would have a $1$ for that bit; those who are indifferent would have a $0$. If it turns out that ``not liking action" is an important attribute, one of the bits could be deployed in that capacity.
We also want a compact representation of each movie. The number of movies is about thirty times fewer than the number of viewers, so we allow for continuous parameters here --- one \textit{attribute weight} corresponding to each attribute bit. One can think of each weight as representing how strongly a movie satisfies the quality of the corresponding bit; e.g., if a movie has a ``likes action'' weight of $+0.5$, viewers with a $1$ in their ``likes action'' bit would be expected to rate this movie $0.5$ stars higher than others who have $0$ in that bit.

We use the attribute bits and attribute weights to explain viewers' ratings of movies. But before we do so, we make some adjustments to the raw data to remove things that are relatively trivial. Suppose we have a matrix in which entry $(v,m)$ contains the rating that viewer $v$ gave to movie $m$ (or is blank if viewer $v$ has not rated movie $m$). First we compute the average of each column (each movie) and subtract it out. This is equivalent to having one extra attribute bit that is always $1$ and a corresponding extra attribute weight representing the overall ``goodness'' of the movie. From the resulting matrix we then subtract out the average of each row (viewer). This accounts for the fact that some viewers may simply be more harsh than others with their ratings. Let $r_{vm}$ denote entry $(v,m)$ of our reduced ratings matrix. The prize winners also preprocessed their data in a similar fashion to give their prediction models a ``baseline'' from which to improve \cite{koren2009bellkor,piotte2009pragmatic,toscher2009bigchaos}.

\begin{figure}
    \centering
    \includegraphics[width=\figwidth]{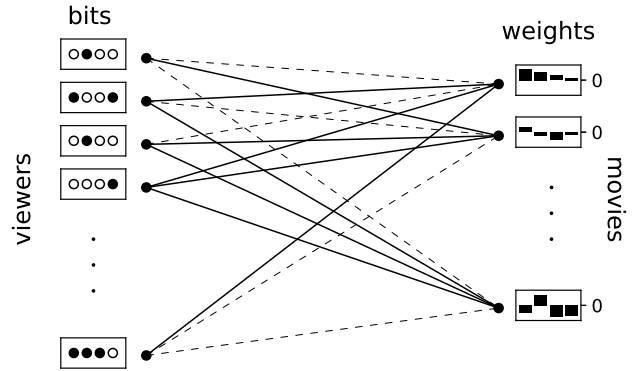}
    \caption[Graphical schematic of BAR]{Schematic of \new{the binary attribute representation (BAR)} as a bipartite graph. Each node on the left represents a viewer and each node on the right represents a movie. Dark edges indicate that the given viewer has rated the corresponding movie, dashed edges indicate the opposite. Each viewer $v$ has a vector $b_v$ of \textit{attribute bits} and each movie $m$ has a vector $w_m$ of \textit{attribute weights}. The objective is to find a set of $b_v$ and $w_m$ such that $b_v\cdot w_m$ is as close as possible to $v$'s rating of $m$ for all $(v,m)$ for which there is a rating.
    }
    \label{fig:network}
\end{figure}

Figure \ref{fig:network} illustrates our model\new{, the binary attribute representation (BAR),} as a bipartite graph. Nodes on the left are viewers, nodes on the right are movies, and the solid edges represent viewer-movie pairs where the viewer has rated the movie. Let $b_v$ be the vector of attribute bits for viewer $v$ and let $w_m$ be the vector of attribute weights for movie $m$. Our goal is to minimize the root mean square error:
\begin{equation}
(\mbox{RMSE})^2=\frac{1}{N}\sum_{v,m} \left(r_{vm}-b_v\cdot w_m\right)^2,
\label{eq:rmse}
\end{equation}
where $N$ is the total number of ratings, and the sum is over all pairs $(v,m)$ where viewer $v$ has rated movie $m$ --- i.e., all solid edges of Figure \ref{fig:network}.

\new{\subsection{Optimization strategy}}
The global optimization problem just described is difficult because each variable is involved in many different constraints --- each $w_m$ is needed for many different viewers, each $b_v$ will be applied to many movies. The fact that each $b_v$ must take discrete values adds an extra wrinkle. Gradient-based optimization is not even applicable because the loss landscape is discontinuous. Sigmoid activation functions and the more popular rectified linear unit \cite{hara2015relu} offer continuous alternatives, \new{but even then the loss landscape would be littered with local minima,} so we take a completely different approach.

First, we define constraints that encode solutions to the optimization problem. Using the divide-and-concur \cite{gravel2008divide} strategy this can be done so that projections to the constraints can be computed efficiently. ``Replicas" are created for each variable --- one for each constraint it is involved in --- and the applicable constraints are imposed with a single projection. Another projection ensures that the replicas are consistent, or concur. Then we use the relaxed reflect-reflect (RRR) algorithm to solve the divide-and-concur system.

\new{For each summand in Eqn. \eqref{eq:rmse} we create replicas of the bit vector and weight vector involved.}
Let $b_{vm}$ be the replica of $b_v$ that is used to explain $v$'s rating of movie $m$, and let $w_{vm}$ be the corresponding replica of $w_m$.
\new{Referring back to Figure \ref{fig:network},} this means that every edge $(v,m)$ now has its own bit vector and weight vector.
We could require $r_{vm}=b_{vm}\cdot w_{vm}$, but since the ratings were originally given as integers from 1 to 5 and the rules of the competition permit non-integer predictions, we allow for rounding error:
\begin{equation}
    |r_{vm}-b_{vm}\cdot w_{vm}|<0.5.
    \label{eq:withreplicas}
\end{equation}
Satisfying inequality \eqref{eq:withreplicas} is a simple bilinear constraint if we allow $b_{vm}$ to be continuous. But we still want our bit vectors to be discrete, so we create one additional replica of the bit vector for each viewer, $c_v$, and impose the discreteness on this replica:
\begin{equation}
    c_v\in\{0,1\}^{\times d}\;,
    \label{eq:discretebit}
\end{equation}
where $d$ is the number of attributes.

Having divided the problem into a collection of easy tasks, we must now make the replicas concur with each other:
\begin{align}
    \forall v,m&:\,b_{vm}=c_v\label{eq:bitconcur}\;,\\
    \forall v_1,v_2,m&:\,w_{v_1m}=w_{v_2m}\;.\label{eq:weightconcur}
\end{align}
For compactness of notation, let us concatenate all $w_{vm}$, $b_{vm}$, and $c_v$ into a single vector and call it $x$. We ultimately want the $b$'s and $c$'s to be discrete, but for now we allow them to be real-valued so that $x$ is a point in a high-dimensional continuous Euclidean space. Let $A$ be the set of $x$ in this space that satisfy \eqref{eq:withreplicas} and \eqref{eq:discretebit}. Let $x^A$ denote the projection of $x$ to $A$. That is, $x^A$ is the point in $A$ that minimizes
\begin{equation}
\sum_{v,m}\left(\|w_{vm}-w^A_{vm}\|^2+\|b_{vm}-b^A_{vm}\|^2\right)+\sum_v \|c_v-c^A_v\|^2.
\label{eq:metric}
\end{equation}
Finding $x^A$ is not difficult because each term in the sum can be handled separately. We have already mentioned that satisfying inequality \eqref{eq:withreplicas} is handled by the projection to a bilinear constraint \cite{elser2021learning,bauschke2022projections} for each $(v,m)$ for which $v$ has rated $m$. If $v$ has not rated $m$ then there is no constraint to satisfy, so we can just set $b^A_{vm}=b_{vm}$ and $w^A_{vm}=w_{vm}$ as this is trivially distance minimizing. The projection to \eqref{eq:discretebit} is the operation of rounding a continuous vector to the nearest vector of bits.

The projections for (\ref{eq:bitconcur}, \ref{eq:weightconcur}) are also straightforward. Let $B$ be the set of $x$ satisfying (\ref{eq:bitconcur}, \ref{eq:weightconcur}) and let $x^B$ denote the projection of a given $x$ to $B$. 
Computing $x^B$ involves nothing more than averaging all of the $w_{vm}$ for each $m$ and all of the $b_{vm}$ and $c_v$ for each $v$.

\new{The search process starts with a random initialization of $x$. Initializing the weight vector replicas with a distribution symmetric about $0$ is natural, given that we have subtracted off the rating averages. There is no definite minimum or maximum value a weight can take, but the range of the original ratings from $1$ to $5$ combined with the fact that up to $d$ attributes can contribute to a given rating leads us to sample uniformly from $[-2/d,+2/d)$. As for the replicas of the bit vectors, we tried random values from $[0,1)$ at first, but we found that led to an unproductive search in which the bits quickly moved toward $0$ or $1$ based purely on their random initial values and stayed there. The algorithm was stuck trying to tune the weight vectors to fit a fixed set of randomly generated bit vectors. There are several ways to avoid this kind of trap in the RRR method \cite{deyo2022avoiding}, but in this case the quickest was to initialize with values from $[0,1/2)$. That way the bits start closer to being ``off'' and the algorithm decides which ones to turn on for each viewer at the same time as it tunes the weights for each movie.

From this initialization we iterate} with the RRR rule \cite{elser2021learning}
\begin{equation}
    x\to \left(1-\frac{\beta}{2}\right)x+\frac{\beta}{2}R_B(R_A(x))\;,
    \label{eq:rrr}
\end{equation}
where 
\begin{equation}
    R_{A,B}(x)=2x^{A,B}-x
\end{equation}
are reflectors in the two constraints. One can check that if $x$ is a fixed point of \eqref{eq:rrr}, then $x^A$ is an element of $A\cap B$; that is, $x^A$ simultaneously satisfies (\ref{eq:withreplicas}-\ref{eq:weightconcur}) and thus is a solution to the global problem.
The parameter $\beta>0$ is analogous to a time step. When all the constraints are convex, $\beta=1$ gives the fastest convergence. Smaller values of $\beta$ are known to improve performance in nonconvex problems like ours. We use $\beta=0.5$. 

\new{There is no reason to expect a global solution to exist for a noisy data set like this one, but at any iteration we can take the values of $c_v$ from $x^A$ as a guess of the viewers' attribute bits and the concurred $w_{vm}$ values from $x^B$ as the movies' attribute weights, and use these to compute the model's RMSE. Figure \ref{fig:error} shows how $x$ and the RMSE evolve. The change $\Delta x$ from one iteration to the next is roughly a measure of how close sets $A$ and $B$ are in the vicinity of $x$. For the first few hundred iterations, the RMSE and $\Delta x$ decline together. If we were expecting exact solutions, then we would expect this trend to continue, with $\Delta x=0$ for a solution. However, we are not expecting exact solutions, so $\Delta x$ will never reach $0$, and at some point its continued diminution is counterproductive. The calculation of RMSE only involves some of the elements of $x$ after all, so it does not have to be correlated with a distance $\Delta x$ that involves every element. Unexpected spikes in RMSE happen occasionally --- there is a small one in the figure just after $250$ iterations --- and if they cause the RMSE to rise above $1.1$ then we reinitialize $x$ and continue iterating. After 1000 iterations, we use whichever iteration achieved the lowest RMSE (not the lowest $\Delta x$) as our ``solution.''
\begin{figure}
    \centering
    \includegraphics[width=\figwidth]{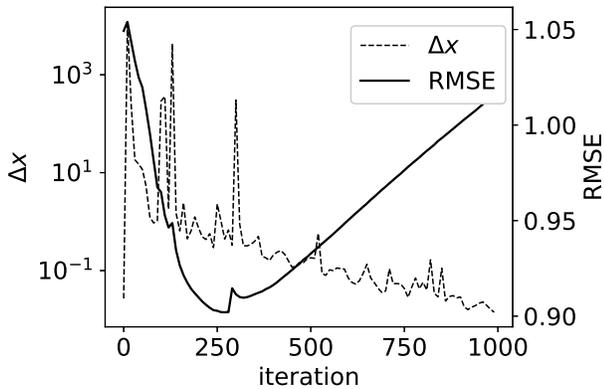}
    \caption[Time series of iteration and prediction error]{Time series of the amount by which $x$ changes in each iteration, $\Delta x$, and the RMSE of the model.}
    \label{fig:error}
\end{figure}
}

The divide-and-concur approach not only makes the computations easy to implement, it also makes them easy to parallelize.
Every viewer-movie pair can be handled independently for \eqref{eq:withreplicas}; every viewer independently for \eqref{eq:discretebit} and \eqref{eq:bitconcur}; and every movie independently for \eqref{eq:weightconcur}. Since each $b_{vm}$, $c_v$, and $w_{vm}$ is a vector with as many elements as there are bits, constraints (\ref{eq:discretebit}-\ref{eq:weightconcur}) can also be parallelized by handling each bit independently. We take advantage of these facts to speed up our implementation when running on multiple cores.

We can make this algorithm more efficient by discarding the replicas $b_{vm}$ and $w_{vm}$ for which $v$ has not rated $m$. This is a significant speedup when the matrix of ratings $r_{vm}$ is sparse. However, this changes the number of terms in the first sum of the metric \eqref{eq:metric}, and we find that we need to adjust the relative weight of the two sums to make this expedited algorithm run effectively. Even with such adjustments, it does not seem to reach quite as low an RMSE as the standard version. Also, our heuristic for selecting a subset of the movies for training (see Section \ref{subsec:selection} below) tends to favor movies with more views, which reduces the sparsity advantage anyway. So we will stick with the standard algorithm for the results presented in this manuscript but we acknowledge that there is a faster version which, with more careful adjustments to the metric, might achieve the same results.

\new{\subsection{Training data}}
\label{subsec:selection}
In the interest of further efficiency, we do not need to train on the entire data set. Some movies deserve more attention than others. To decide which movies deserve the most attention, consider a naive model that ``predicts'' every viewer will give every movie its average rating. This is a useful baseline model for comparing with the less trivial approaches to the Netflix challenge \cite{feuerverger2012statistical}. The squared error of this method is the sum of squares of every actual rating minus the corresponding movie average. If we rank the movies by how much they contribute to this error, we find that the first $100$ of the $17770$ movies account for $15\%$ of the squared error, and the first $500$ account for $44\%$ of the error. 

We hypothesize that a set of attribute bits that can explain viewers' ratings on a few hundred of the most contentious (influential) movies will do a fine job on the rest of the movies. So we train on such a subset to obtain attribute bits ($b_v$) for all viewers. Then, it is a straightforward exercise to minimize
\begin{equation}
\sum_{v,m} (r_{vm}-b_v\cdot w_m)^2
\end{equation}
with respect to $w_m$ for all the movies (a linear system for each movie).

A repository with code for finding attribute bits and weights is available at https://github.com/seandeyo/BAR.


\section{Experiments}
\label{sec:exp}
Our goal is to demonstrate that \new{BAR} can explain the Netflix data set through a small number of discrete bits for each viewer, in a representation that is easy to interpret, and that these parameters can be learned from a small subset of the movies. To that end, in this section we 
\begin{itemize}
    \item vary the size of the small subset,
    \item vary the number of bits each viewers gets, and
    \item illustrate how to interpret the parameters the algorithm learns.
\end{itemize}


\subsection{Training subset}
Our approach, as described in the previous section, is to train only on a subset of the movies that contribute the most to the squared error of the ``every movie gets its average rating'' method. The hypothesis (hope) is that if a small subset of movies covers a sizeable portion of this baseline error, then the attribute bits inferred to explain the viewers' ratings in the subset will be just as good at explaining the viewers' ratings of the rest of the movies.

\begin{figure}
    \centering
    \includegraphics[width=\figwidth]{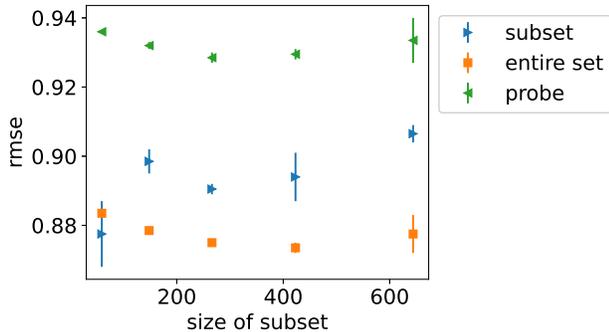}
    \caption[RMSE vs.\ size of training subset]{Root mean square error (RMSE) for several sizes of training subset, using eight attributes. To select the subsets, we rank the movies by how much they contribute to error of the ``every movie gets its average rating'' method. We take movies from the top of that ranking until the subset covers a certain fraction of the total error: $10\%$ (59 movies), $20\%$ (148 movies), $30\%$ (266 movies), $40\%$ (423 movies), $50\%$ (644 movies). When we apply the attribute bits learned from training on the subset to the rest of the movies the resulting RMSE is usually lower than it was for the subset alone, supporting the notion that most of the hard work is in a relatively small number of movies.}
    \label{fig:subsets}
\end{figure}

The natural question is how much of the baseline error our movie subset should cover in order to achieve optimal performance. To answer this question we use five subsets, containing $10\%$, $20\%$, $30\%$, $40\%$, or $50\%$ of the baseline error (always starting with the movies that contribute the most). We train on each subset, then use the inferred bits to find weights for the entire training set. The results are given in Figure \ref{fig:subsets}.  
Note that the RMSE on the whole data set is generally lower than on the subset. This supports our hypothesis that most of the hard work lies in a relatively small number of movies; the rest are easier to explain. Indeed, training on the subset that covers $50\%$ of the baseline error appears to offer no advantage over the subset covering $20\%$. \new{We trained on each subset twice, with a different random initial $x$ in Eqn. \eqref{eq:rrr} each time, to see how consistent the results were. The points in Figure \ref{fig:subsets} indicate the averages over these two trials, with the error bars reflecting how different the two trials were. There is variation of around $0.01$ in the subset RMSE in some cases, but the entire-set RMSE is more consistent.}

We also provide the RMSE on the ``probe,'' which is a special subset of the data. To evaluate the competitors, Netflix used a hidden set of ``qualifying data'' --- a set of movie-viewer pairs for which a rating existed but that rating was not in the training data. The competitors had to predict the ratings of the qualifying data, and they would only find out their performance after making their submission, so having a subset of the training data --- the probe --- identified as statistically similar to the qualifying data was helpful for them to get a sense of how well their methods were working. The qualifying data, and thus the probe set, are not statistically identical to the overall training data: The point of a movie recommendation system is to use the viewers' older ratings to predict their newer ones, so the qualifying data and probe skew toward more recent ratings. They are also almost uniform among the viewers (the training set is not uniform in this regard), and are slightly less nonuniform among the movies \cite{feuerverger2012statistical}. 

As our model makes no attempt to address these differences, and indeed makes no use at all of the recency or frequency of ratings (which the prize winners found to be useful in making predictions \cite{piotte2009pragmatic}), we do not expect it to perform particularly well on the probe. We are more concerned with the RMSE on the training data, and on demonstrating that it can be achieved with a small amount of information stored in a way that is easy to interpret. Indeed, our RMSE is several percent higher on the probe than on the whole data set; nonetheless, we include it for the sake of comparison. The prize winner achieved an RMSE of $0.856$ on the test set, though perhaps a more reasonable benchmark would be the typical RMSE of one of the ingredients in the blend, which mostly fell in the range of $0.88-0.90$ \cite{koren2009bellkor,piotte2009pragmatic,toscher2009bigchaos}. Typical RBMs achieve around $0.92$ or $0.91$ with 100 or 500 features \cite{salakhutdinov2007rbm}, or $0.90$ with 1000 features \cite{zhou2008collaborative}. Modified RBMs employed by the prize winners, with additional parameters to account for the dates on which ratings were made, improve to around $0.895$ \cite{koren2009bellkor}.

\subsection{Number of attributes}
\begin{figure}
    \centering
    \includegraphics[width=\figwidth]{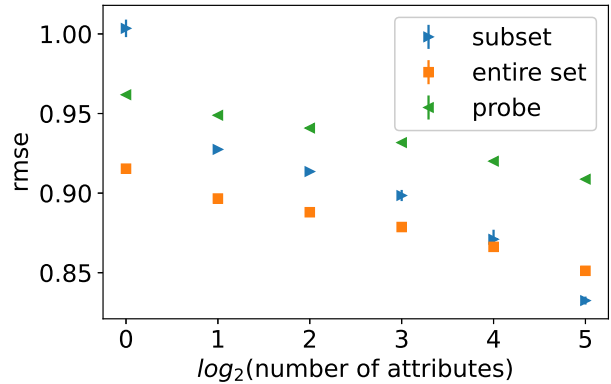}
    \caption[RMSE vs.\ number of attributes]{Root mean square error (RMSE) for several numbers of attributes, training on a 148-movie subset of the data. The RMSE on the entire data set (and on the probe set) decreases uniformly with the number of attribute bits, each doubling of bits reducing the RMSE by about $0.01$. As in Figure \ref{fig:subsets}, the RMSE is higher on the training subset --- except with $32$ bits, which suggest an overfitting of the subset.}
    \label{fig:attributes}
\end{figure}

How many attribute bits per viewer are necessary to reach our benchmark RMSE range? To find out, we take one of our subsets, the one with $148$ movies that covered $20\%$ of the baseline error, and train with 1, 2, 4, 8, 16, and 32 bits. The results are given in Figure \ref{fig:attributes}.
Naturally, the RMSE decreases as the number of bits increases. The decrease is roughly logarithmic in the number of bits, with a drop of about $0.01$ in 
the entire-set or probe RMSE every time the number of bits doubles. The pattern of RMSE being higher on the subset than on the entire set begins to reverse around $16$ or $32$ bits, suggesting to us that we may be overfitting the subset at that point. The number of bits at which this behavior occurs is also the number of bits at which the entire-set RMSE drops below the test-set RMSE of the grand prize winners, and the probe RMSE is comparable to that of the RBMs. Thus, we conclude that this data set can be explained with reasonable accuracy with 16 or perhaps 32 bits per viewer, as opposed to the hundreds of features that the RBMs used \cite{salakhutdinov2007rbm,zhou2008collaborative}.

\subsection{Interpreting a solution}

\begin{figure}
    \centering
    \includegraphics[width=\figwidth]{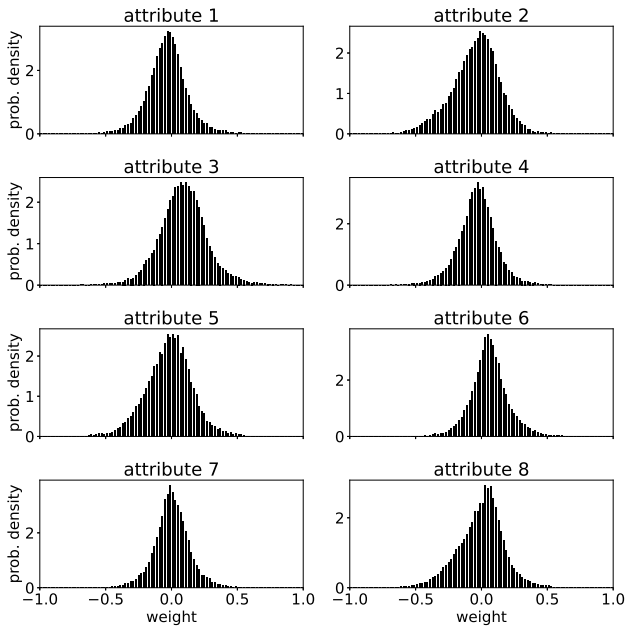}
    \caption[Example histograms of attribute weights]{Histograms of attribute weights in one solution trained on the $148$-movie subset.}
    \label{fig:weights}
\end{figure}
If we want to look deeper to have a more subjective description of what each attribute represents, we can look at some movies with similar weights. Figure \ref{fig:weights} shows histograms of the weights for each attribute in the $8$-attribute model trained on the $148$-movie subset. They are generally centered near but not precisely at zero, and they often have a skew to one side or the other. The inferred weights are not the same from one trial to the next, of course --- the bit numbers are subject to permutation symmetry, some attributes in one trial might be more or less linear combinations of other attributes from another trial, etc. 

Nonetheless, patterns are usually clear if we look at the movies with especially high or low values for a given weight. Table \ref{tab:weights1} does this for attribute number 1 from Figure \ref{fig:weights}.
\begin{table}[t]
    \centering
    \begin{tabular}{l|r}
        Title & Weight\\
        \hline
        \textit{Fight Club} & -0.67\\
        \textit{Gone in 60 Seconds} & -0.62\\
        \textit{Con Air} & -0.50\\
        \textit{The Rock} & -0.44\\
        \textit{Dirty Dancing} & -0.42\\
        \textit{Lethal Weapon 4} & -0.39\\
        \textit{Armageddon} & -0.38\\
        \textit{Pulp Fiction} & -0.37\\
        \textit{American Pie} & -0.36\\
        \textit{The Fast and the Furious} & -0.35\\
        \hline
        \textit{Sweet Home Alabama} & 0.20\\
        \textit{The Wedding Planner} & 0.21\\
        \textit{How to Lose a Guy in 10 Days} & 0.22\\
        \textit{Maid in Manhattan} & 0.22\\
        \textit{Erin Brockovich} & 0.22\\
        \textit{Sister Act} & 0.28\\
        \textit{Something's Gotta Give} & 0.28\\
        \textit{Two Weeks Notice} & 0.29\\
        \textit{Miss Congeniality} & 0.58\\
        \textit{My Big Fat Greek Wedding} & 0.58\\
        \hline
    \end{tabular}
    \caption[Titles with extreme weights for attribute 1]{A selection of movies and their weights for attribute number 1 in a solution trained on the $148$-movie subset (the same solution as in Figure \ref{fig:weights}), highlighting the films with especially high or low weights.}
    \label{tab:weights1}
\end{table}
The films with positive scores here are arguably women-centered, while those with negative scores have largely male casts and, one imagines, largely young male audiences. We might call this the ``not a young male'' attribute. Table \ref{tab:weights2} does the same for attribute number 2. 
\begin{table}[t]
    \centering
    \begin{tabular}{l|r}
        Title & Weight\\
        \hline
        \textit{Lost in Translation} & -1.24\\
        \textit{Punch-Drunk Love} & -0.93\\
        \textit{The Royal Tenenbaums} & -0.91\\
        \textit{Adaptation} & -0.83\\
        \textit{Being John Malkovich} & -0.80\\
        \textit{Eternal Sunshine of the Spotless Mind} & -0.76\\
        \textit{Rushmore} & -0.76\\
        \textit{The Big Lebowski} & -0.75\\
        \textit{Annie Hall} & -0.72\\
        \textit{The Life Aquatic with Steve Zissou} & -0.70\\
        \hline
        \textit{National Treasure} & 0.40\\
        \textit{Two Weeks Notice} & 0.42\\
        \textit{Men in Black II} & 0.43\\
        \textit{Coyote Ugly} & 0.52\\
        \textit{Maid in Manhattan} & 0.53\\
        \textit{How to Lose a Guy in 10 Days} & 0.54\\
        \textit{Sweet Home Alabama} & 0.57\\
        \textit{Pearl Harbor} & 0.60\\
        \textit{The Wedding Planner} & 0.60\\
        \textit{Miss Congeniality} & 0.81\\
        \hline
    \end{tabular}
    \caption[Titles with extreme weights for attribute 2]{A selection of movies and their weights for attribute number 2 in a solution trained on the $148$-movie subset (the same solution as in Figure \ref{fig:weights}), highlighting the films with especially high or low weights.}
    \label{tab:weights2}
\end{table}
On the negative end we see films that might be described as quirky. Many of these have narratives without a definitive or tidy resolution, and a director or writer with a distinct style like Wes Anderson (\textit{The Royal Tenenbaums}, \textit{Rushmore}, \textit{The Life Aquatic with Steve Zissou}) or Charlie Kaufman (\textit{Adaptation}, \textit{Being John Malkovich}, \textit{Eternal Sunshine of the Spotless Mind}).
This might be the ``does not like quirkiness'' attribute. Indeed, on the positive end we have more conventional blockbusters in which the protagonists overcome the central conflict and get their tidy resolution. Some are romantic comedies (\textit{Sweet Home Alabama}, \textit{The Wedding Planner}), some are action flicks (\textit{Men in Black II}, \textit{National Treasure} --- and others like \textit{The Fast and the Furious} and \textit{Gone in 60 Seconds} narrowly missed the top ten with weights of $0.38$). The highest scorer, \textit{Miss Congeniality}, is both. In any case, these are not the kind of movies that leave the viewer pondering how to interpret the story, at least not in the same way as the negative-weight group.

Our descriptions of what these attributes represent are certainly subjective, but the point is that the process does not have to be difficult or mysterious. We can clearly see the qualitative differences between the movies with high scores and low scores, and we could list all the viewers that care about this difference by checking which ones have a 1 in the corresponding attribute bit.

\section{Conclusion}

The awarding of the Netflix prize was a key moment in the early development of machine learning. The objective of the competition was not to make a simple, interpretable classification system, but to get the lowest RMSE by whatever means necessary, and the winners of the prize did just that. We can hardly blame them --- a million dollars is a strong motivator --- but there are many applications of data classification that would benefit from more simplicity and interpretability under the hood. The difficulty is that more complicated models seem to be more successful, and the trend in the years since the Netflix competition has been to embrace the complexity. Many of the models used in the ``blend'' that won the Netflix prize had a suite of meta-parameters that had to be finely tuned, as did the blending process itself, but this amalgamation looks tame in comparison with the deep neural networks in use today. We have returned to the Netflix data set to demonstrate that an approach committed to transparency and interpretability can achieve an RMSE similar to one of the blend ingredients using only a few bits of information for each viewer. We hope this exercise shows that accuracy need not always come at the expense of simplicity and transparency.

\bibliography{refs,commonrefs}

\providecommand{\noopsort}[1]{}\providecommand{\singleletter}[1]{#1}%
\begin{thebibliography}{11}%
\makeatletter
\providecommand \@ifxundefined [1]{%
 \@ifx{#1\undefined}
}%
\providecommand \@ifnum [1]{%
 \ifnum #1\expandafter \@firstoftwo
 \else \expandafter \@secondoftwo
 \fi
}%
\providecommand \@ifx [1]{%
 \ifx #1\expandafter \@firstoftwo
 \else \expandafter \@secondoftwo
 \fi
}%
\providecommand \natexlab [1]{#1}%
\providecommand \enquote  [1]{``#1''}%
\providecommand \bibnamefont  [1]{#1}%
\providecommand \bibfnamefont [1]{#1}%
\providecommand \citenamefont [1]{#1}%
\providecommand \href@noop [0]{\@secondoftwo}%
\providecommand \href [0]{\begingroup \@sanitize@url \@href}%
\providecommand \@href[1]{\@@startlink{#1}\@@href}%
\providecommand \@@href[1]{\endgroup#1\@@endlink}%
\providecommand \@sanitize@url [0]{\catcode `\\12\catcode `\$12\catcode
  `\&12\catcode `\#12\catcode `\^12\catcode `\_12\catcode `\%12\relax}%
\providecommand \@@startlink[1]{}%
\providecommand \@@endlink[0]{}%
\providecommand \url  [0]{\begingroup\@sanitize@url \@url }%
\providecommand \@url [1]{\endgroup\@href {#1}{\urlprefix }}%
\providecommand \urlprefix  [0]{URL }%
\providecommand \Eprint [0]{\href }%
\providecommand \doibase [0]{https://doi.org/}%
\providecommand \selectlanguage [0]{\@gobble}%
\providecommand \bibinfo  [0]{\@secondoftwo}%
\providecommand \bibfield  [0]{\@secondoftwo}%
\providecommand \translation [1]{[#1]}%
\providecommand \BibitemOpen [0]{}%
\providecommand \bibitemStop [0]{}%
\providecommand \bibitemNoStop [0]{.\EOS\space}%
\providecommand \EOS [0]{\spacefactor3000\relax}%
\providecommand \BibitemShut  [1]{\csname bibitem#1\endcsname}%
\let\auto@bib@innerbib\@empty
\bibitem [{\citenamefont {Feuerverger}\ \emph {et~al.}(2012)\citenamefont
  {Feuerverger}, \citenamefont {He},\ and\ \citenamefont
  {Katri}}]{feuerverger2012statistical}%
  \BibitemOpen
  \bibfield  {author} {\bibinfo {author} {\bibfnamefont {A.}~\bibnamefont
  {Feuerverger}}, \bibinfo {author} {\bibfnamefont {Y.}~\bibnamefont {He}},\
  and\ \bibinfo {author} {\bibfnamefont {S.}~\bibnamefont {Katri}},\
  }\href@noop {} {\bibfield  {journal} {\bibinfo  {journal} {Statistical
  Science}\ }\textbf {\bibinfo {volume} {27}},\ \bibinfo {pages} {202}
  (\bibinfo {year} {2012})}\BibitemShut {NoStop}%
\bibitem [{\citenamefont {Koren}(2009)}]{koren2009bellkor}%
  \BibitemOpen
  \bibfield  {author} {\bibinfo {author} {\bibfnamefont {Y.}~\bibnamefont
  {Koren}},\ }\href@noop {} {\bibfield  {journal} {\bibinfo  {journal} {Netflix
  prize documentation}\ } (\bibinfo {year} {2009})}\BibitemShut {NoStop}%
\bibitem [{\citenamefont {Piotte}\ and\ \citenamefont
  {Chabbert}(2009)}]{piotte2009pragmatic}%
  \BibitemOpen
  \bibfield  {author} {\bibinfo {author} {\bibfnamefont {M.}~\bibnamefont
  {Piotte}}\ and\ \bibinfo {author} {\bibfnamefont {M.}~\bibnamefont
  {Chabbert}},\ }\href@noop {} {\bibfield  {journal} {\bibinfo  {journal}
  {Netflix prize documentation}\ } (\bibinfo {year} {2009})}\BibitemShut
  {NoStop}%
\bibitem [{\citenamefont {T{\"o}scher}\ \emph {et~al.}(2009)\citenamefont
  {T{\"o}scher}, \citenamefont {Jahrer},\ and\ \citenamefont
  {Bell}}]{toscher2009bigchaos}%
  \BibitemOpen
  \bibfield  {author} {\bibinfo {author} {\bibfnamefont {A.}~\bibnamefont
  {T{\"o}scher}}, \bibinfo {author} {\bibfnamefont {M.}~\bibnamefont
  {Jahrer}},\ and\ \bibinfo {author} {\bibfnamefont {R.~M.}\ \bibnamefont
  {Bell}},\ }\href@noop {} {\bibfield  {journal} {\bibinfo  {journal} {Netflix
  prize documentation}\ } (\bibinfo {year} {2009})}\BibitemShut {NoStop}%
\bibitem [{\citenamefont {Salakhutdinov}\ \emph {et~al.}(2007)\citenamefont
  {Salakhutdinov}, \citenamefont {Mnih},\ and\ \citenamefont
  {Hinton}}]{salakhutdinov2007rbm}%
  \BibitemOpen
  \bibfield  {author} {\bibinfo {author} {\bibfnamefont {R.}~\bibnamefont
  {Salakhutdinov}}, \bibinfo {author} {\bibfnamefont {A.}~\bibnamefont
  {Mnih}},\ and\ \bibinfo {author} {\bibfnamefont {G.}~\bibnamefont {Hinton}},\
  }in\ \href@noop {} {\emph {\bibinfo {booktitle} {Proceedings of the 24th
  international conference on Machine learning}}}\ (\bibinfo {year} {2007})\
  pp.\ \bibinfo {pages} {791--798}\BibitemShut {NoStop}%
\bibitem [{\citenamefont {Hara}\ \emph {et~al.}(2015)\citenamefont {Hara},
  \citenamefont {Saito},\ and\ \citenamefont {Shouno}}]{hara2015relu}%
  \BibitemOpen
  \bibfield  {author} {\bibinfo {author} {\bibfnamefont {K.}~\bibnamefont
  {Hara}}, \bibinfo {author} {\bibfnamefont {D.}~\bibnamefont {Saito}},\ and\
  \bibinfo {author} {\bibfnamefont {H.}~\bibnamefont {Shouno}},\ }in\
  \href@noop {} {\emph {\bibinfo {booktitle} {2015 international joint
  conference on neural networks (IJCNN)}}}\ (\bibinfo {organization} {IEEE},\
  \bibinfo {year} {2015})\ pp.\ \bibinfo {pages} {1--8}\BibitemShut {NoStop}%
\bibitem [{\citenamefont {Gravel}\ and\ \citenamefont
  {Elser}(2008)}]{gravel2008divide}%
  \BibitemOpen
  \bibfield  {author} {\bibinfo {author} {\bibfnamefont {S.}~\bibnamefont
  {Gravel}}\ and\ \bibinfo {author} {\bibfnamefont {V.}~\bibnamefont {Elser}},\
  }\href@noop {} {\bibfield  {journal} {\bibinfo  {journal} {Physical Review
  E}\ }\textbf {\bibinfo {volume} {78}},\ \bibinfo {pages} {036706} (\bibinfo
  {year} {2008})}\BibitemShut {NoStop}%
\bibitem [{\citenamefont {Elser}(2021)}]{elser2021learning}%
  \BibitemOpen
  \bibfield  {author} {\bibinfo {author} {\bibfnamefont {V.}~\bibnamefont
  {Elser}},\ }\href@noop {} {\bibfield  {journal} {\bibinfo  {journal} {Fixed
  Point Theory and Algorithms for Sciences and Engineering}\ }\textbf {\bibinfo
  {volume} {2021}},\ \bibinfo {pages} {1} (\bibinfo {year} {2021})}\BibitemShut
  {NoStop}%
\bibitem [{\citenamefont {Bauschke}\ \emph {et~al.}(2022)\citenamefont
  {Bauschke}, \citenamefont {Lal},\ and\ \citenamefont
  {Wang}}]{bauschke2022projections}%
  \BibitemOpen
  \bibfield  {author} {\bibinfo {author} {\bibfnamefont {H.~H.}\ \bibnamefont
  {Bauschke}}, \bibinfo {author} {\bibfnamefont {M.~K.}\ \bibnamefont {Lal}},\
  and\ \bibinfo {author} {\bibfnamefont {X.}~\bibnamefont {Wang}},\ }\href@noop
  {} {\bibfield  {journal} {\bibinfo  {journal} {Journal of Global
  Optimization}\ ,\ \bibinfo {pages} {1}} (\bibinfo {year} {2022})}\BibitemShut
  {NoStop}%
\bibitem [{\citenamefont {Deyo}\ and\ \citenamefont
  {Elser}(2022)}]{deyo2022avoiding}%
  \BibitemOpen
  \bibfield  {author} {\bibinfo {author} {\bibfnamefont {S.}~\bibnamefont
  {Deyo}}\ and\ \bibinfo {author} {\bibfnamefont {V.}~\bibnamefont {Elser}},\
  }\href@noop {} {\bibfield  {journal} {\bibinfo  {journal} {Journal of Applied
  and Numerical Optimization}\ }\textbf {\bibinfo {volume} {4}},\ \bibinfo
  {pages} {143} (\bibinfo {year} {2022})}\BibitemShut {NoStop}%
\bibitem [{\citenamefont {Zhou}\ \emph {et~al.}(2008)\citenamefont {Zhou},
  \citenamefont {Wilkinson}, \citenamefont {Schreiber},\ and\ \citenamefont
  {Pan}}]{zhou2008collaborative}%
  \BibitemOpen
  \bibfield  {author} {\bibinfo {author} {\bibfnamefont {Y.}~\bibnamefont
  {Zhou}}, \bibinfo {author} {\bibfnamefont {D.}~\bibnamefont {Wilkinson}},
  \bibinfo {author} {\bibfnamefont {R.}~\bibnamefont {Schreiber}},\ and\
  \bibinfo {author} {\bibfnamefont {R.}~\bibnamefont {Pan}},\ }in\ \href@noop
  {} {\emph {\bibinfo {booktitle} {Algorithmic Aspects in Information and
  Management: 4th International Conference, AAIM 2008, Shanghai, China, June
  23-25, 2008. Proceedings 4}}}\ (\bibinfo {organization} {Springer},\ \bibinfo
  {year} {2008})\ pp.\ \bibinfo {pages} {337--348}\BibitemShut {NoStop}%
\end{thebibliography}%

\end{document}